\documentclass[letterpaper, 10 pt, conference]{ieeeconf}  

\IEEEoverridecommandlockouts                              

\usepackage{graphics} 
\usepackage{epsfig} 
\usepackage{mathptmx} 
\usepackage{times} 
\usepackage{amsmath} 
\usepackage{amssymb}  
\usepackage[hidelinks]{hyperref}   
\usepackage[nameinlink,capitalise,noabbrev]{cleveref} 
\usepackage{tabularx}   
\usepackage{bm}
\usepackage{siunitx}
\usepackage{booktabs}
\usepackage{stfloats}
\usepackage{caption}
\title{\LARGE \bf
Real-Time Arm Posture Estimation For Robot-aided Physiotherapy via RGB-D–Robot Sensor Fusion 
}

\author{Yuxuan Hu$^{1}$*, Aditya Srivastava, Luka Peternel$^{1}$, Laura Marchal-Crespo$^{1}$ and J. Micah Prendergast$^{1}$
\thanks{$^{1}$ Department of Cognitive Robotics,
        TU Delft, Delft, The Netherlands
        Corresponding author: {\tt\small yuxuanhu@tudelft.nl}}%
}

\begin{document}

\maketitle
\thispagestyle{empty}
\pagestyle{empty}

\begin{abstract}
 Real-time estimation of human arm posture and movement can support biomechanical feedback during robot-aided physiotherapy. This paper presents a markerless sensor-fusion framework that combines RGB-D-derived arm-segment directions with end-effector robot-derived hand pose and twist through a biomechanics-constrained extended Kalman filter (EKF). We evaluated seven-degree-of-freedom arm joint angle and joint velocity estimation in 10 participants performing a physical human–robot interaction task, using motion-capture-based biomechanical inverse kinematics as the reference. With participant-specific hand-to-tool (grip) transformations fitted using motion-capture reference, pooled median absolute errors were \SI{5.8}{\degree} for joint angles and \SI{5.0}{\degree}/s for joint velocities. Matched EKF ablations showed that fusion reduced median joint angle error relative to robot-only estimation by \SI{27.0}{\percent}. The framework produced estimates at an average of \SI{58.5}{\hertz} on a laptop. Comparison with assumed grip modeling showed that distal-joint angle estimates depended strongly on the hand-to-end-effector transformation. A proof-of-concept demonstration illustrated the use of estimated arm posture to generate real-time, model-derived shoulder-strain feedback during robot-aided physiotherapy movements. These findings characterize the benefits and limitations of camera–robot fusion for robot-aided physiotherapy. The code and dataset are publicly available on \href{https://github.com/TUD-HRC/robot_rgbd_arm_pose_estimation}{\textcolor{blue}{GitHub}}.
\end{abstract}

\section{INTRODUCTION}
In robot-aided physiotherapy, real-time estimates of arm joint angles and velocities can inform biomechanics-based feedback and guide robotic assistance~\cite{prendergast2021biomechanics}. Accurate estimation of human arm movement in a robotic setup, however, remains a challenge~\cite{GP_velocity}. While solving for human arm pose in real-time is straightforward with exoskeletons that have sufficient degrees of freedom~\cite{exo_harmony, armin}, a full-arm exoskeleton requires adjustments to the user's arm dimensions and alignment of its mechanical joints with the corresponding anatomical joints, in addition to its cost, which limits its accessibility. In contrast, end-effector robots can guide movement through a handle or attachment point, avoiding the need for joint-by-joint alignment. However, this simpler physical interface introduces a sensing challenge for estimating human arm pose. 

A 7-degree-of-freedom human arm is kinematically redundant with respect to its 6D hand pose. So a unique solution for the human arm pose can not be determined solely from the pose of a robot end-effector connected to the human hand/arm. Previous robot-based approaches have addressed this ambiguity through sensor attachment~\cite{7DoFIMU}, ergonomic constraints~\cite{robot_7DoF_arm}, probabilistic posture estimation~\cite{Yazdani}, or learned human posture preferences~\cite{GP_velocity}. To bridge the lack of sensing,~\cite{GP_velocity} and~\cite{Yazdani} assume knowledge of the human starting posture. This assumption doesn't hold true during the proposed walk-up-and-use robot interaction.  

Vision-based feedback offers another path for the estimation of real-time arm pose, with the most reliable solution being multi-camera optical motion capture with reflective markers~\cite{mocap_review}. Despite the benefits of this solution, it requires a complex setup with multiple cameras, as well as donning and doffing of dedicated suits or a set of reflective markers. Over the last two decades, vision-based human pose tracking without reflective markers has undergone substantial improvement, with pipelines such as MediaPipe~\cite{mediapipe} and OpenPose~\cite{Openpose} providing real-time detection of human body landmarks from single-view videos. Combining image-based detections with depth measurements enables estimation of the landmarks' 3D positions~\cite{mediapipe_on_realsense, kalman_filter_depth}. Landmark positions can subsequently be used to estimate joint angles and to support real-time biomechanical analysis~\cite{cegliaRealtimeAccurateOpen2025a, bio_pose}. Yet, these single-camera approaches often suffer from large errors in estimating axial limb rotations, such as forearm supination~\cite{cegliaRealtimeAccurateOpen2025a}. In addition, single-camera approaches are highly susceptible to self-occlusion from the human body or occlusion from the robot linkages~\cite{hwangNovelEndEffectorRobot2020}.

\begin{figure}[t!]
    \centering
    \includegraphics[width=\linewidth]{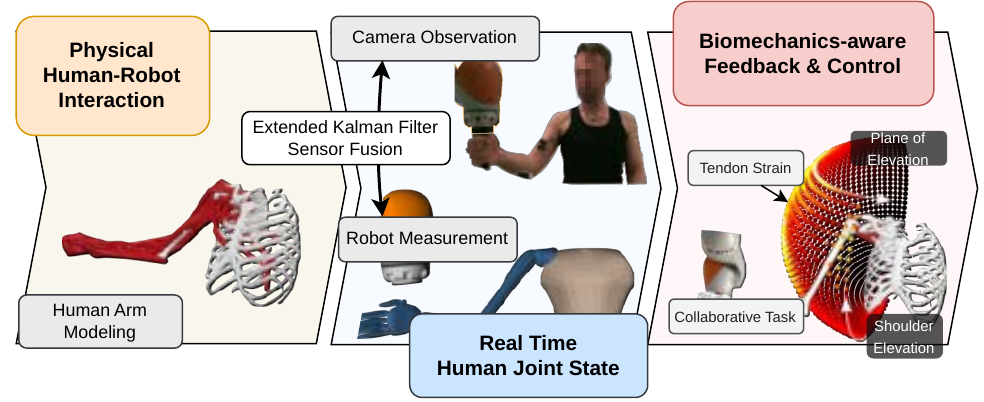}
    \caption{Overview of the proposed framework for real-time pose estimation of a 7-DoF human arm by an extended Kalman filter with robot and vision measurements. During robot-aided physiotherapy, this would allow us to estimate pose-inferred physical states such as tendon strain, providing real-time visual feedback. }
    \label{fig:abstract}
\end{figure}

To address these limitations and to support the objective of ``walk-up-and-use'' physically assistive robots, this work seeks to estimate arm joint angles and velocities during interaction with an end-effector robot without requiring the user to wear markers or sensors. In this direction, previous studies on camera–robot fusion have shown the potential to augment both measurements and address their own limitations. Hwang \textit{et al.} proposed a direct analytical method of using robot-coupled wrist positions to supplement 4-DoF arm pose estimation from a Kinect system~\cite{hwangNovelEndEffectorRobot2020}, which addressed camera occlusion to some extent. But their work assumed no wrist movement, which limits its application in rehabilitation. Yazdani \textit{et al.} fused OpenPose detections with robot measurements using a particle filter and showed robustness over initial human postures and camera occlusion~\cite{yazdaniOcclusionRobustMultiSensoryPosture2022}. Yet this work provided no evaluation of joint velocity accuracy and computational timing, both of which are essential for real-time control and feedback in robot-aided physiotherapy. In addition, the modeling of the hand-to-end-effector transformation, which describes how the user grips the tool, has not been well discussed in either work. 

\begin{figure*}[t!]
  \centering
  \vspace{2mm}
  \includegraphics[width=0.9\textwidth]{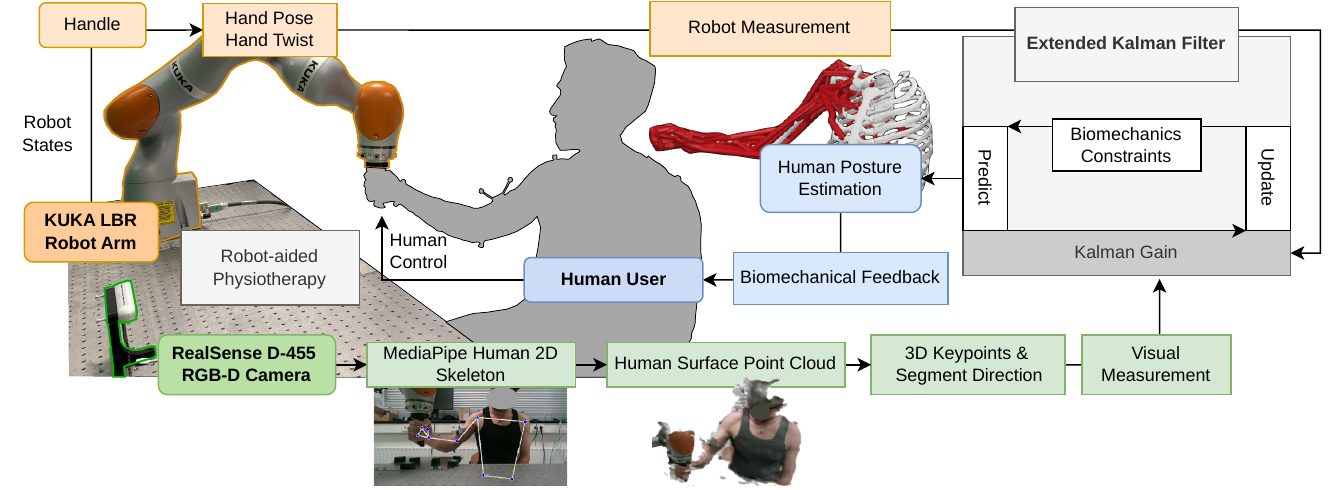}
  \caption{Overview of the proposed arm pose estimation system. An annotated image of the hardware setup is shown on the left. Data flow for the EKF pipeline is shown in the block diagram. The orange route depicts how data from the robot arm is processed and utilized as a high-frequency measurement input to the EKF. The green route shows the data flow for computer-vision measurements. The EKF uses both measurement sources and corrects the pose estimate in real time. The pose estimation provides biomechanical information in the right arm, which we display to the human user as feedback in a proof-of-concept robot-aided physiotherapy application.}
  \label{fig:overview}
  \vspace{-2mm}
\end{figure*} 

We address these gaps with a biomechanics-constrained extended Kalman filter (EKF) that fuses RGB-D-measured arm-segment and robot-derived hand pose and twist estimates, thereby estimating seven human joint angles and their corresponding velocities. The contributions of this work include: (1) a high-frequency fusion framework for an RGB-D camera and an end-effector robot, and its systematic validation with 10 participants in a physical human-robot interaction task, with EKF ablations to assess each measurement source; (2) the investigation of the sensitivity of grip modeling by comparing an assumed hand-to-end-effector transformation shared across participants with participant-specific transformations fitted to the reference data. 

The key use case for the proposed system is to provide visual feedback and control an end-effector robotic arm during robot-aided physiotherapy based on fast real-time estimation of human posture (\cref{fig:abstract}). Previous work conducted by our group includes robotic guidance of shoulder movement based on estimated strain in the shoulder tendons~\cite{prendergast2021biomechanics, belli2025shared}, yet such a system is limited by the accuracy of human pose estimation. Following the aforementioned systematic validation, as a proof of concept, we demonstrate how the estimated posture can update a model-derived shoulder strain map in real time, providing visual feedback during a robot-aided physiotherapy exercise. 

\section{Methods}\label{sec:approach}

\subsection{System Setup and Kinematic Modeling}
Our system consists of a KUKA LBR iiwa 7 robot (KUKA AG, Augsburg, Germany) and an Intel RealSense D-455 RGB-D camera (Intel Corp., Santa Clara, CA, USA). RGB images and point clouds from the camera, as well as robot joint states, are communicated at \SI{15}{\hertz} and \SI{1}{\kilo\hertz}, respectively, to the host laptop running ROS 2 Jazzy. Extrinsic calibration is performed to align the robot base frame and camera frame. See an annotated image of the hardware setup in \cref{fig:overview}.

\begin{figure}[t!]
  \centering
  \vspace{2mm}
  \includegraphics[width=\linewidth]{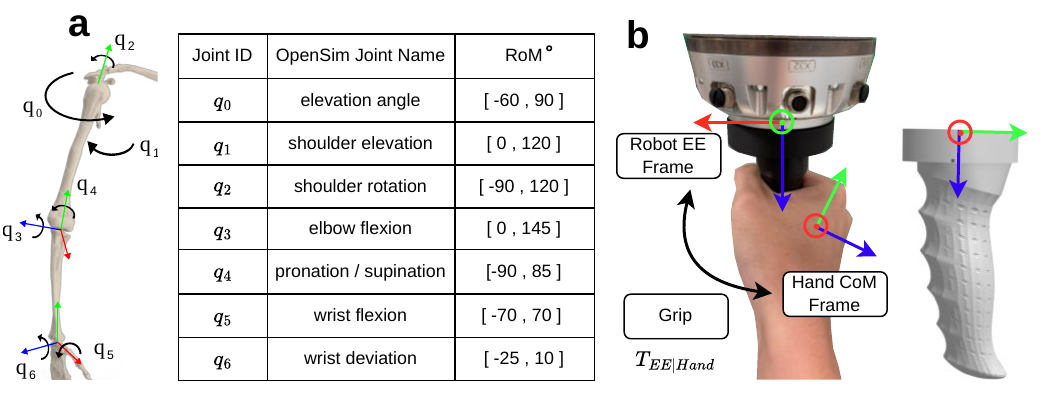}
  \caption{a. Illustration of arm kinematics defined for this study. The table defines IDs and naming conventions in the OpenSim convention for individual joints, along with the biomechanics-based constraints (in degrees) used in the proposed system. b. Illustration of the static transformation from the robot EE frame to the hand CoM frame, describing the grip, as well as the side view CAD model of the ergonomic handle (adapted from~\cite{ergonomic_handle}). }\label{fig: arm_kinematics}
  \vspace{-2mm}
\end{figure}

We model the unilateral human arm with 7 degrees of freedom (DoFs) (\cref{fig: arm_kinematics}.a) according to biomechanical conventions \cite{bio_convention}, which allows the reconstruction of human arm movements using OpenSim \cite{opensim}. We assume the user's shoulder joint is stabilized and no glenohumeral translation occurs during arm movements. When the user is sitting upright in the chair, we assume the user's trunk is parallel to the camera's image plane. 

To capture biomechanically reliable ground truth of human arm movement for validation, we set up the system in a motion-capture environment equipped with six OptiTrack (NaturalPoint Inc., Oregon, USA) Prime$^X$ cameras and motion capture software Motive: Body 2.2.0. Eleven reflective markers were placed on each participant's right arm and torso, adapted from the experimental setup of~\cite{opensim_arm_model}. Before the experiment, the participants were guided into a static neutral pose for scaling of the human skeleton in OpenSim. Arm segment lengths were then extracted from the scaled model and incorporated into the unified robot description format (URDF) used by the posture estimator. 

The human interacts with the robot through a custom-made ergonomic end effector (EE) attached to the robot's flange to encourage a consistent gripping pose. The handle design is adapted from the publicly available design in~\cite{ergonomic_handle}. We assume that the grip remains fixed throughout each task and represent it by a static transformation from the EE frame to the human hand frame $T_{EE\vert Hand} = (R, t)$ defined in CAD, where
\begin{equation} \label{eq:hand_transform}
    R = R_x(87^\circ)\,R_y(66^\circ)\,R_z(-80^\circ), \quad t = [0,\ 0,\ 50]^T~\text{mm}
\end{equation}
This transformation is used as the universal grip model for all participants (see \cref{fig: arm_kinematics}.b). 

Because a participant's actual grip may differ from the modeled one, we also examined performance under a reference-calibrated grip model. To this end, we estimated a participant-specific transformation directly from the motion-capture reference. At each sampling time, we compared the hand orientation reconstructed from the reference joint angles with that obtained from the robot end-effector orientation using the universal grip transformation. 

The grip correction is defined as the mean relative rotation applied to $T_{EE\vert Hand}$ that minimizes the error between the two hand orientations throughout the entire recording. The translational component was left unchanged because the handle design was considered sufficient to reliably constrain the hand position. We evaluated posture estimation using both the corrected and universal grip transformations to assess sensitivity to grip-model mismatch.

\subsection{Arm Pose Estimation via Extended Kalman Filtering}
The filter estimates the state of the 7-DoF arm, described by the joint angles and corresponding joint velocities,
\begin{equation}
    \bm{x} = \begin{bmatrix} \bm{q}^{\!\top} & \dot{\bm{q}}^{\!\top} \end{bmatrix}^{\!\top} \in \mathbb{R}^{14},
\end{equation}
where $\bm{q}, \dot{\bm{q}} \in \mathbb{R}^{7}$ are the joint angles and angular velocities, expressed relative to a shoulder-anchored base frame. 

The prediction uses a standard constant-velocity model with white-noise acceleration. After every prediction, the joint limits provided by the biomechanics model (see~\cref{fig: arm_kinematics}.a) are enforced by truncating the probability density function~\cite{EKF_truncation}, thereby projecting the estimates into the feasible anatomical RoM while consistently updating covariance $\bm{P}$.

The EKF runs at up to \SI{60}{\hertz}, incorporating robot-state measurements received at \SI{1}{\kilo\hertz} and camera measurements at \SI{15}{\hertz}. Measurements are incorporated through standard EKF corrections using the residuals, Jacobians, and noise covariances defined below.

\subsubsection{Robot-Derived Hand Pose and Twist}
We transform the rapidly updating robot end-effector pose and twist into measurements of the hand pose $(\bm{p}^{m}, \bm{R}^{m})$ and twist $\bm{\nu}^{m} \in \mathbb{R}^6$ in the shoulder frame through a static transformation describing the grip (see \cref{eq:hand_transform}). Comparing these with the hand pose and twist predictions by the forward kinematics of the human arm yields the residuals
\begin{equation}
    \bm{r}_p = \bm{p}^{m} - \bm{p}(\bm{q}), \quad
    \bm{r}_R = \operatorname{Log}\!\big(\bm{R}^{m} \bm{R}(\bm{q})^{\!\top}\big), \quad
    \bm{r}_\nu = \bm{\nu}^{m} - \bm{J}(\bm{q})\,\dot{\bm{q}}.
\end{equation}

The measurement Jacobian $\bm{H}$ of each correction is defined by the kinematic hand Jacobian of the arm:
\begin{equation}
    \bm{H}_p = \begin{bmatrix} \bm{J}_v(\bm{q}) & \bm{0} \end{bmatrix}, \quad
    \bm{H}_R = \begin{bmatrix} \bm{J}_\omega(\bm{q}) & \bm{0} \end{bmatrix}, \quad
    \bm{H}_\nu = \begin{bmatrix} \bm{0} & \bm{J}(\bm{q}) \end{bmatrix}.
    \label{eq:hand_H}
\end{equation}

The pose residuals act on the joint angles $\bm{q}$ through separate linear ($\bm{J}_v$) and angular ($\bm{J}_\omega$) Jacobians, while the twist residual acts on the joint angular velocities $\dot{\bm{q}}$ through the full arm Jacobian. Robot-derived hand measurements alone do not uniquely determine the redundant arm configuration, motivating additional vision-based segment measurements.


\subsubsection{Vision-Based Segment Directions}
We obtain 2D skeletal keypoints (shoulder, elbow, wrist) in the RGB image using MediaPipe~\cite{mediapipe}, one of many open-source real-time 2D human pose detectors. Intel RealSense cameras provide an aligned point cloud to the RGB images, giving us the 3D location of every point on the observed human surface, which we then use to estimate 3D skeletal keypoints with an empirically determined surface-to-joint-center offset of \SI{5}{\centi\meter} medially and \SI{5}{\centi\meter} upward for the shoulder, and a depth of \SI{4}{\centi\meter} for the elbow and \SI{2}{\centi\meter} for the wrist.

We determine the right shoulder's location in the world frame from the first 5 seconds since system start. At each frame, we compute 3D unit directions $\bm{u}^{m}_{u}$ and $\bm{u}^{m}_{f}$ of the upper arm $\bm{u}$ (shoulder$\to$elbow) and forearm $\bm{f}$ (elbow$\to$wrist) segments. For the upper arm direction, we subtract an empirical residual bias of \SI{5}{\degree} from both the elevation angle and the shoulder elevation. The empirical surface-to-joint offset and upper-arm bias were selected by visually inspecting the alignment between the reconstructed arm and point cloud in a recording of one author, excluded from the evaluation dataset. 

A frame is rejected if a measured segment length lies outside \SIrange{60}{125}{\percent} of the participant's measured limb length. During invalid vision measurements, a zero-velocity pseudo-measurement on the elbow swivel keeps the null space motion from drifting. 

Residuals of the vision-based measurements are computed and applied sequentially for the two segments $j \in \{u, f\}$: 
\begin{equation}
    \bm{r}_{j} = \bm{u}^{m}_{j} - \bm{u}_{j}(\bm{q}), \quad
    \bm{H}_{j} = \begin{bmatrix} \frac{\bm{I}_3 - \bm{u}_{j}\bm{u}_{j}^{\!\top}}{\lVert\bm{s}_{j}\rVert}
    \big(\bm{J}_{v,e} - \bm{J}_{v,s}\big) & \bm{0} \end{bmatrix},
    \label{eq:bone_dir}
\end{equation}
where $\bm{u}_{j}(\bm{q})$ is the unit direction of the segment vector $\bm{s}_{j}$ from forward kinematics, and $\bm{J}_{v,s}, \bm{J}_{v,e}$ are the linear Jacobians of the segment start and end frames (shoulder and elbow for the upper arm, elbow and wrist for the forearm). 

All reported results were obtained with the same EKF parameters, with isotropic measurement noise covariances $\bm{R} = \sigma^2\bm{I}$. Standard deviations $\sigma$ for each measurement can be found in the released repository. Camera measurements are applied to the current EKF state without temporal alignment or delay compensation. The most recent keypoint set is reused at each cycle until replaced, provided its depth-frame timestamp is no more than \SI{0.2}{\second} earlier than the current robot sample. Each vision measurement is therefore reused an average of 3 times, reducing its effective $\sigma$, which then acts as a tuned weight between robot and camera measurements rather than a calibrated noise level. 


\section{Evaluation Design}
\begin{figure}[t!]
    \centering
    \includegraphics[width=\linewidth]{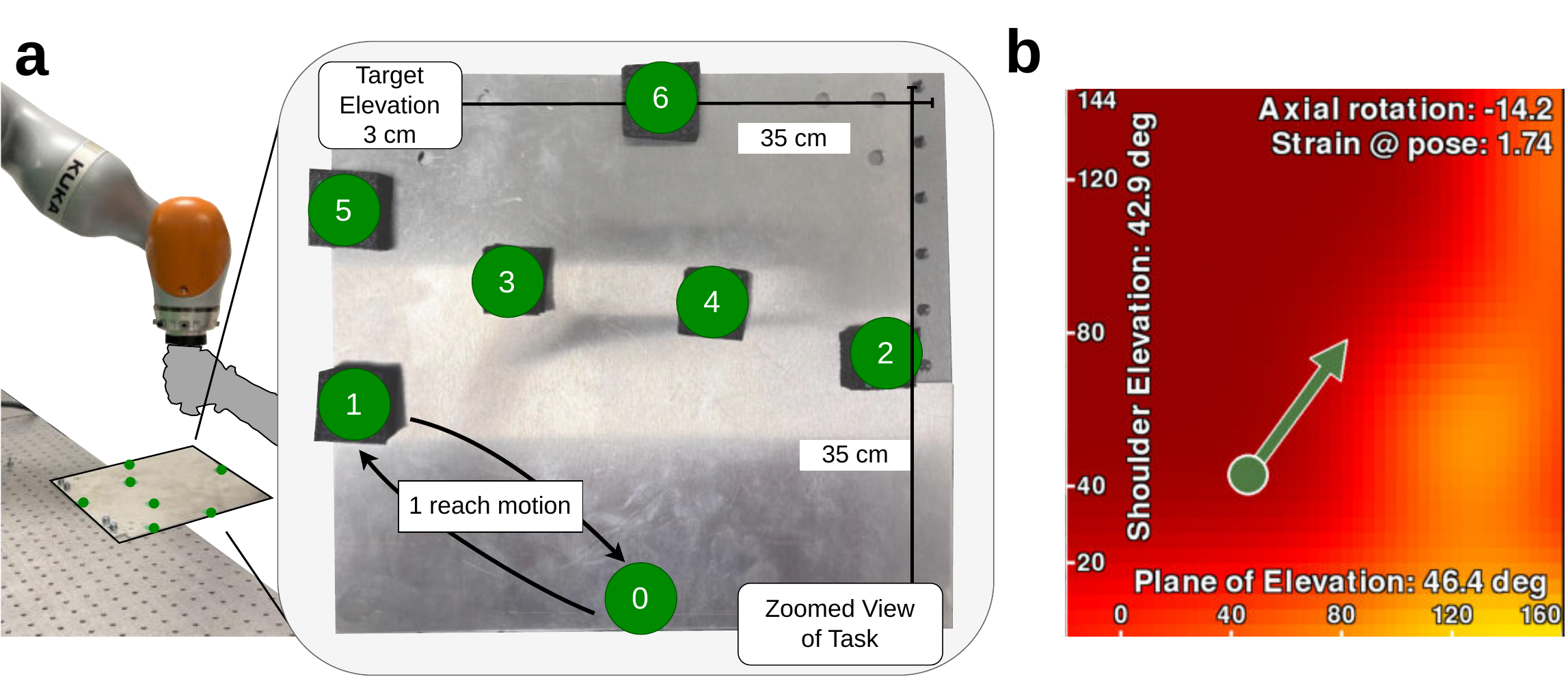}
    \caption{a. Annotated illustration of the tabletop reaching task used in the experiment. The targets, labeled 1--6, are \SI{3}{\cm}-tall sponge blocks attached to an aluminum board. An example reaching trajectory is also shown on the target board. b. User interface in the demonstration to provide visual feedback on shoulder pose and estimated strain. Fonts and cursor are enlarged for illustration purposes.}
    \label{fig:experiment_setup}
\end{figure}

\subsection{Experimental Setup}
We conducted the experiment with 10 healthy participants (5 female and 5 male) aged 24 to 35 years. The participants were all right-handed, students/researchers in robotics, and were between \SI{155}{\cm} and \SI{192}{\cm} in height. All participants provided written consent to participate in the study. The procedures followed the protocol approved by the Human Research Ethics Committee of TU Delft (HREC number 5778).

We implemented Cartesian impedance control on the KUKA robot, with a translational stiffness of 10~$\mathrm{N/m}$ and a rotational stiffness of 0.5~$\mathrm{N\cdot m/rad}$ at the EE, to initiate a relaxed physical interaction with non-trivial resistance. The participants were instructed to perform a reaching task while holding the robot EE. Six \SI{3}{\cm} sponge targets were placed within a horizontal planar surface of \SI{35}{\cm} by \SI{35}{\cm} (see the arrangement in \cref{fig:experiment_setup}). Participants were instructed to use the robot EE to lightly tap each target, starting from a table-level starting position. After becoming familiar with the robot, the participants were instructed to complete two trials. In trial A, the participants reached targets 1--6 in sequence and repeated the sequence three times. In trial B, participants reached each target three consecutive times before proceeding to the next target, repeating this procedure for targets 1--6. The participants were instructed to complete the tasks at whatever pace they found comfortable, with no time limits or feedback. 

\subsection{Data Analysis}
Raw data from the experimental setup were recorded as ROS~2 bag files, and the results were computed in real time by replaying the recordings at their original speed. All 10 participants finished the prescribed trajectories, yielding a total of 360 reaching motions in 20 recordings. Estimates from the proposed system and all ablation configurations (described later) were compared against motion-capture-based ground truth. The reference joint angles were computed from the marker data by OpenSim inverse kinematics at \SI{240}{\hertz}. We define the posture estimation error as the absolute difference in joint angles between the estimated pose and the motion-capture reference.

To assess whether the proposed data fusion improves pose estimation accuracy, we conducted an ablation study using three configurations of the proposed EKF: (1) robot-only measurements, (2) vision-only measurements, and (3) fused measurements. All configurations used the same EKF parameters and initial arm pose, with the shoulder in a neutral position, the upper arm resting alongside the torso, the elbow flexed to $90^\circ$, and the forearm and wrist in neutral positions. As an additional baseline, we also include the pose estimate from an online inverse-kinematics (IK) solver using the KDLKinematicsPlugin provided by MoveIt 2 \cite{moveit2}, with the same initial pose as above. All modules in the ablation studies used the participant-specific corrected grip transformation $T_{EE\vert Hand}$. Finally, to assess sensitivity to grip modeling, we ran the fused EKF module using the universal grip transformation.
\subsection{Posture-based shoulder strain feedback}

To demonstrate how the estimated arm posture can support real-time feedback in robot-aided physiotherapy, we integrate the proposed method with the biomechanics-based shoulder strain map used in our previous work~\cite{belli2025shared}. In that work, estimated strain in the shoulder tendons was precomputed in OpenSim as a function of the three shoulder DoFs, and saved locally as 3D maps for each tendon. By stacking these maps, we obtain a 3D map that describes the peak strain achieved by any tendon, with position on the map determined by shoulder pose; see the right panel of \cref{fig:abstract} for an illustration of this concept. 

One generalized biomechanical simulation is used for calculation, and the resulting interactive strain map is shown to one of the authors. We display on the screen the user's position and velocity on the strain map as a green cursor and arrow; see \cref{fig:experiment_setup} (b). The user should use this visual feedback to reduce shoulder strain during trajectory execution. 

As a proof-of-concept demonstration, the user, with a healthy shoulder RoM, was instructed to slowly perform a predefined physiotherapy movement: from a neutral arm pose (upper arm close to torso, elbow flexed at \SI{90}{\degree}) to a lifted arm pose (arm fully extended, hand raised at shoulder level). The user performed this trajectory twice: once with visual feedback on their position in the strain map, once without. In both cases, the user attempted to avoid creating high strain on the shoulder while reaching for the same task-space target with their hand. We compare movement history to see if the user successfully used the visual feedback to avoid high shoulder strain.

\section{Results}

\subsection{Pose Estimation Accuracy}
\begin{figure}[t!]
    \centering
    \vspace{2mm}
    \includegraphics[width=\linewidth]{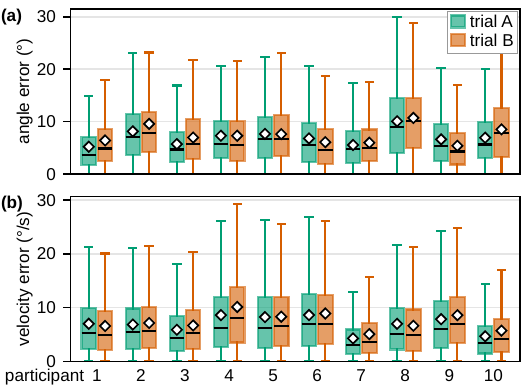}
    \caption{Distribution of the absolute joint angle (a) and joint velocity (b) errors in the proposed method (all DoFs combined), per participant (n = 10) and trial: trial A in green, trial B in orange. Boxes show the interquartile range, horizontal lines indicate the median, whiskers extend to 1.5 times the interquartile range, and white diamonds indicate the mean.}
    \label{fig:subject_trial}
    \vspace{-2mm}
\end{figure}

\begin{table*}[!h]
\vspace*{3mm}
\centering
\footnotesize
\setlength{\tabcolsep}{2pt}
\caption{Summary of statistics for joint angle and joint velocity estimation accuracy validation. Median, Q3, and $P_{95}$ are of the absolute error. We also report the range of motion (for joint angles and velocities) achieved by all participants as the central 95\% of the entire dataset for this joint. Static biases in joint angles are reported as [minimum, maximum] accross participants. Errors are expressed as $x(y)$: $x$ with the participant-specific corrected grip pose, and $y$ with the universal grip pose. }
\label{tab:ekf_error_angles_velocity}
\begin{tabular}{lccccc|cccc}
\toprule
 & \multicolumn{5}{c|}{a. Joint angle ($^\circ$)} & \multicolumn{4}{c}{b. Joint velocity ($^\circ$/s)}\\
\cmidrule(lr){2-6}\cmidrule(lr){7-10}
Degree of freedom & Median & Q3 & $P_{95}$ & Bias & Range & Median & Q3 & $P_{95}$ & Range\\
\midrule
Elevation angle      & 6.2 (7.1) & 11.2 (12.2) & 20.4 (21.0) & $[-7.8, +11.4]$ ($[-5.9, +13.4]$) & $[9.9, 57.7]$ & 6.8 (6.8) & 12.3 (12.4) & 21.6 (22.1) & $[-21.1, 22.4]$ \\
Shoulder elevation   & 3.6 (3.7) & 6.8 (6.6) & 12.9 (12.5) & $[-10.6, +1.8]$ ($[-9.3, +2.1]$) & $[31.5, 67.2]$ & 3.1 (3.3) & 5.9 (6.3) & 11.5 (11.9) & $[-17.8, 19.4]$ \\
Shoulder rotation    & 6.4 (6.3) & 9.8 (10.6) & 15.3 (16.8) & $[-11.0, +8.1]$ ($[-12.8, +8.5]$) & $[-27.8, 8.3]$ & 3.8 (3.9) & 7.2 (7.3) & 13.8 (13.9) & $[-16.2, 15.8]$ \\
Elbow flexion        & 4.9 (5.1) & 8.8 (8.7) & 15.8 (16.2) & $[-6.0, +1.8]$ ($[-6.9, +3.1]$) & $[42.5, 93.1]$ & 6.9 (7.1) & 12.7 (13.1) & 24.5 (25.5) & $[-32.6, 39.3]$ \\
Pronation supination & 5.9 (10.2) & 10.2 (14.7) & 18.4 (28.1) & $[-8.1, +9.6]$ ($[-28.6, +12.3]$) & $[-29.9, 12.8]$ & 4.7 (4.5) & 8.7 (8.1) & 16.9 (15.3) & $[-12.2, 11.3]$ \\
Wrist flexion        & 7.2 (14.6) & 12.0 (22.2) & 21.2 (37.4) & $[-2.6, +12.7]$ ($[-20.3, +31.0]$) & $[-69.1, -16.7]$ & 5.4 (6.2) & 10.2 (11.8) & 19.8 (23.0) & $[-19.9, 19.9]$ \\
Wrist deviation      & 7.6 (8.0) & 12.6 (12.8) & 19.3 (20.2) & $[-3.7, +10.7]$ ($[-15.4, +10.7]$) & $[-21.5, 2.5]$ & 6.3 (6.6) & 11.7 (11.8) & 21.9 (21.5) & $[-16.5, 13.6]$ \\
\midrule
All DoFs combined    & 5.8 (7.0) & 10.2 (12.5) & 18.2 (23.8) & $[-2.2, +2.3]$ ($[-5.9, +1.6]$) & --- & 5.0 (5.2) & 9.7 (9.9) & 19.5 (20.1) & --- \\
\bottomrule
\end{tabular}

\end{table*}

We first report the detailed temporal performance of the proposed system. On a laptop equipped with an Intel Core Ultra 7 155H processor and 32 GB of RAM, we replayed all 20 ROS 2 bags in real time through the complete pipeline. One fused EKF cycle took a median of \SI{0.24}{\milli\second} (99th percentile: \SI{0.41}{\milli\second}) at an average of \SI{58.5}{\hertz}. Robot measurements reached the estimate at a median of \SI{0.31}{\milli\second} after their publication. Camera information reached the estimate with a median of \SI{186}{\milli\second} ($P_{95}$ \SI{206}{\milli\second}) after capturing the RGB frame supplying the landmarks, or \SI{123}{\milli\second} ($P_{95}$ \SI{144}{\milli\second}) after capture of the associated depth frame, due to the inherent delay between the RGB and depth images. The vision pipeline rejected MediaPipe due to self-occlusion at a median of \SI{19.3}{\percent} (maximum \SI{74.2}{\percent}) of the processed camera frames in trial A, and \SI{19.1}{\percent} (maximum \SI{46.6}{\percent}) in trial B, across participants; both maxima belong to participant 8.

We present the accuracy of the proposed EKF method in estimating joint angles and velocities. In \cref{tab:ekf_error_angles_velocity}, we report the median, third quartile (Q3), and 95th percentile ($P_{95}$) of the absolute errors relative to the motion-capture ground truth for each joint, aggregated across all 10 participants. In addition, we report the achieved range of motion in each joint (in motion-capture reference) by any participant, in the form of the interval containing the central 95\% ($P_{2.5}$ to $P_{97.5}$) in the joint's motion history, combined for all participants.

To assess how much error is induced by the grip pose assumption, we report error terms in the format $x(y)$: $x$ with the participant-specific corrected grip pose, and $y$ with the universal grip pose. We also report the signed static bias in joint angles ($q_{\text{estimated}} - q_{\text{reference}}$) as [minimum, maximum] accross participants to reflect potential modeling error. Relative to the universal grip, the participant-specific grip corrections amounted to a median rotation of \SI{18.8}{\degree} (range \SIrange{11.8}{37.9}{\degree}) across the 20 recordings, and varied most about the hand x-axis, which corresponds to wrist flexion (\SIrange{-27.8}{24.3}{\degree}). The corrections differed less between the two trials of a participant (median \SI{6.6}{\degree}) than between participants (median \SI{26.4}{\degree}), indicating that each participant held the handle consistently, but differently from the others.

We also report, when using corrected grip for each participant and trial, participant-level and trial-level comparisons of joint angle and velocity errors in \cref{fig:subject_trial}. All participants show joint angle errors with median $\leq 10.1^\circ$ and Q3 $\leq 14.5^\circ$, and both trials yielded similar results (\cref{fig:subject_trial}.a). Participant 8 showed slightly higher error rates than the rest, which corresponded to the higher self-occlusion percentage reported earlier. Velocity estimation has similar results across participants (\cref{fig:subject_trial}.b). 

\begin{figure}[t!]
  \centering
  \vspace{2mm}
  \includegraphics[width=\linewidth]{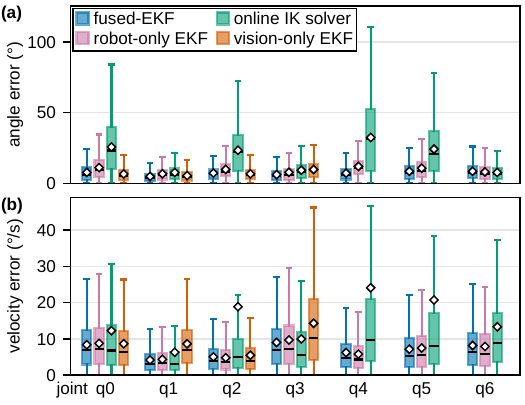}
  \caption{Joint angle (a) and joint velocity (b) estimation errors in the ablation study for the proposed method (labeled as fused-EKF in blue), compared against robot-only EKF in pink, an online IK solver in green, and vision-only EKF in orange, pooled across all participants and both trials. Boxes show the interquartile range, horizontal lines indicate the median, whiskers extend to 1.5 times the interquartile range, and white diamonds indicate the mean.}
  \label{fig: joint_boxplot}
  \vspace{-2mm}
\end{figure}


\begin{figure}[t!]
    \centering
    \includegraphics[width=\linewidth]{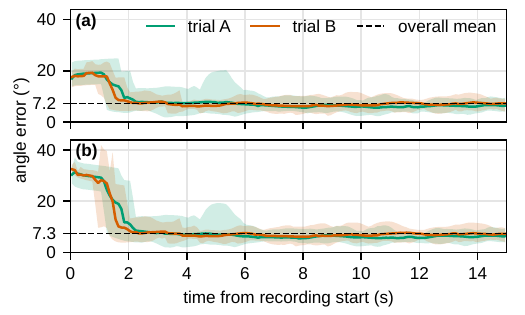}
    \caption{Absolute joint angle error over all DoFs plotted against time, from the start of the recording. Lines show the mean across participants for trial A (green) and trial B (orange). Shaded areas span the minimum to the maximum across participants, and dashed lines show the overall mean joint angle error.}
    \label{fig:convergence}
\end{figure}

We show results of the matched EKF albation in \cref{fig: joint_boxplot}. The robot-only EKF substantially outperforms the online IK solver, reducing the median absolute error by \SI{38.4}{\percent} (median across participants, IQR $[21.4\%, 52.6\%]$). The improvement replicates in 8 of 10 participants and in both trials independently. Adding the camera measurements further improves the estimate: the fused EKF reduces the median error of the robot-only EKF by \SI{27.0}{\percent} (IQR $[15.3\%, 37.7\%]$) and outperforms it in 9 of 10 participants (19 of 20 recordings), see \cref{fig: joint_boxplot}.a. Compared with the vision-only EKF on the four DoFs that it observes ($q_0$--$q_3$), the fused EKF also improves the estimate, reducing the median error by \SI{15.5}{\percent} (IQR $[6.1\%, 26.9\%]$) in all 10 participants. 

For velocity estimation (\cref{fig: joint_boxplot}.b), fused EKF's errors are slightly higher than the robot-only EKF's in 7 of 10 participants. Both estimators improve on their vision-only counterpart in shoulder elevation and in elbow flexion, the fastest-moving joint according to the recorded velocity ranges; in elbow flexion, the fused EKF reduces the median velocity error of the vision-only EKF by \SI{32.0}{\percent} (IQR $[27.7\%, 34.4\%]$), in all 10 participants.

To show convergence time of the fused EKF when supplied with different initial pose, we plot the absolute joint angle error over time, from the beginning of the recording in \cref{fig:convergence}. In (a), we plot the results initialized at the previously described neutral pose, while in (b), we initialize the system with joint angles on each joint's positive limit (moved inward by the same truncation). The error reaches its steady level within about 2\,s from the neutral pose and 2.6\,s from the joint limits. Most of this time passes before the vision pipeline first detects the shoulder (median 1.7\,s), at which point the EKF has no measurement to correct. 

Lastly, we investigate the system's robustness to vision dropouts, which are primarily casued by self-occlusion. In \cref{tab:dropout}, we bin periods with invalid vision measurements by duration, and report their occurrence in the experiments in number of periods (n) and overall time, as well as their effects on the pose estimation. We see that longer dropout periods were associated with higher joint angle errors in the fused EKF, while joint velocity errors remained at a similar level. 

\begin{table}[t]
\centering
\footnotesize
\setlength{\tabcolsep}{2pt}
\caption{Mean ($P_{95}$) absolute error during occurrence of frame rejection, binned by continued duration. Results shown are for all DoFs, participants, and trials combined. $n$: number of dropouts.}
\label{tab:dropout}
\begin{tabular}{lrrcccc}
\toprule
 & & & \multicolumn{2}{c}{Joint angle ($^\circ$)} & \multicolumn{2}{c}{Joint velocity ($^\circ$/s)}\\
\cmidrule(lr){4-5}\cmidrule(lr){6-7}
Dropout & $n$ & Time (s) & Fused & Robot-only & Fused & Robot-only\\
\midrule
None        & ---  & 1164 & 6.5 (16.0) & 9.1 (22.1) & 6.8 (19.0) & 6.9 (19.7)\\
$<1$\,s     & 514  & 123  & 7.2 (17.7) & 9.5 (22.6) & 7.2 (20.3) & 7.3 (20.6)\\
1--2\,s     & 100  & 146  & 8.7 (21.3) & 10.3 (26.0) & 7.3 (21.2) & 7.3 (21.5)\\
2--3\,s     & 36   & 84   & 9.0 (22.2) & 10.2 (27.0) & 7.8 (22.7) & 7.7 (22.4)\\
$\geq 3$\,s & 10   & 45   & 10.1 (24.6) & 12.9 (31.4) & 7.0 (19.0) & 7.0 (18.6)\\
\midrule
All         & ---  & 1562 & 7.0 (17.5) & 9.4 (23.2) & 6.9 (19.5) & 7.0 (20.1)\\
\bottomrule
\end{tabular}
\end{table}



\subsection{Biomechanics-aware Robot-aided Physiotherapy}

\begin{figure*}[t!]
    \centering
    \vspace{2mm}
    \includegraphics[width=\textwidth]{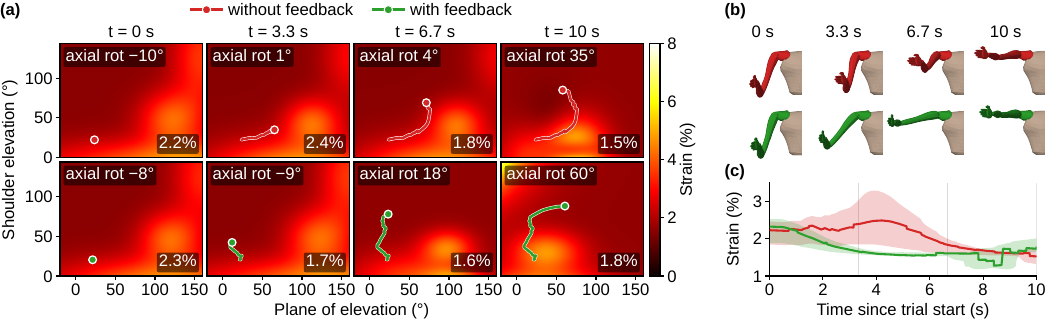}
    \caption{Trajectory comparison between shoulder physiotherapy movements with (green) and without (red) real-time shoulder strain map feedback. a) 2D trajectory history on the strain map at \SI{0}{\second}, \SI{3.3}{\second}, \SI{6.7}{\second} and \SI{10}{\second}. The x-axis denotes the elevation angle, and the y-axis denotes shoulder elevation. The map changes based on the shoulder rotation value and reflects the current 2D slice. b) Estimated arm postures at the aforementioned timestamps, visualized in ROS~2. c) Time series of shoulder strain (\%) for both conditions. Shaded areas show the 5th--95th percentile of strain given the shoulder joint estimation errors.}
    \label{fig:strain_image}
    \vspace{-2mm}
\end{figure*}

The movement history of the user's two attempts (with and without visual feedback) is shown in \cref{fig:strain_image}. In (a), we see the comparison of shoulder joint-space trajectories, visualized in 2D slices of the strain map. When provided with visual feedback of the shoulder position in the strain map, the user effectively took a route with less strain. We see in (b) that in both conditions, the user started and ended in similar postures. But with visual feedback, the user was able to avoid high-strain regions during the motion by performing larger shoulder elevation. 

We additionally plotted the estimated strain over time in (c). For each shoulder pose sample in both attempts, we add the entire distribution of signed joint angle errors derived from \cref{tab:ekf_error_angles_velocity}, assuming a corrected grip transformation for the user. We then calculate the strain distribution given the added error, and plot its $5^{th}$--$95^{th}$ percentile as shaded regions. As expected, we see greater uncertainty in strain due to pose-estimation noise near the high-strain region. 

\section{Discussion}
We presented a sensor fusion system using an RBG-D camera and an end-effector robot that can estimate human arm joint states at around \SI{60}{\hertz}, with accuracy that generally agrees with motion capture ground truth. Our results in the matched EKF ablation showed that combining RGB-D and robot measurements improved joint angle accuracy across participants, when corrected grip transformation is applied. This finding supports combining robot EE and RGB-D measurements to estimate arm posture. The presented errors in joint angles, however, should be interpreted with consideration that the motion capture reference itself contains noise~\cite {upper_body_accuracy}, caused by marker placement, soft-tissue artifacts, and general modeling errors of human limbs. 

For joint velocity, the benefit of fusion was less evident in the matched ablation. We report no apparent improvement in joint velocity estimation going from the robot-only EKF to the fused EKF. We also see that velocity estimation maintains its accuracy during prolonged vision dropouts. One possible explanation is that much of the joint motion during the tested task contributed to hand movement and was therefore reflected in the measured end-effector twist, making it an effective measurement source. The joint velocity ranges in \cref{tab:ekf_error_angles_velocity} provide context for this interpretation. In movements that contain more null-space velocity, for example, a baseball throw, end-effector twist will be less informative about joint velocity. 

Our results show that proper grip calibration was necessary to ensure good tracking of distal joint angles. By using the motion-capture reference to determine participant-specific grip transformations, we see substantial improvement in the accuracy for pronation/supination and wrist flexion. This finding quantifies the importance of accounting for individual differences in grip when estimating distal joint angles. We believe that future work on calibrating the grip transformation using prescribed movement rather than the motion-capture reference would improve the accessibility of this system. 

As an additional finding, the vision dropout analysis suggests that sensor fusion benefits the system's accuracy during short visual interruptions. The fused estimator maintained lower joint angle errors than the robot-only estimator during these interruptions. However, the fused estimator's advantage narrowed as the duration of vision dropout increased. This is consistent with the complementary roles of the two sensors: arm movements in the null space can go undetected by robot measurements until visual measurements resume. 

We acknowledge several limitations to the reported findings: 1) Reported accuracy is limited to a table-level pick-and-place task by the dominant hand. See \cref{tab:ekf_error_angles_velocity} for the range of motion reached by the participants. 2) In the vision pipeline, we rely on empirical parameters that are determined specifically for the hardware setup. In a setup where the camera is differently placed, certain empirical parameters will require recalibration. 3) The participant-specific grip transformations were both fitted and evaluated on the same motion-capture recordings. The improved agreement therefore does not establish accuracy on unseen movements, and the fitted transformations may also compensate for other modeling discrepancies.

The system also shows the following disadvantage: even with participant-specific grip calibration, the 95th percentile of pooled absolute joint angle errors was \SI{18.2}{\degree}. Large errors occurred in elevation angle, wrist flexion, and wrist deviation. We show the danger of large estimation noise in \cref{fig:strain_image}.c, where $P_{95}$ errors in the shoulder joints lead to uncertain prediction of tendon strain, especially around high strain areas. Qualitative inspection of the recordings suggested two separate explanations. Torso rotation about the longitudinal axis could affect estimates of elevation angle because the model assumes the torso faces the camera. Similarly, a change in grip during a task could affect wrist-angle estimates because the model assumes a fixed hand-to-end-effector transformation. Improved modeling of human posture and grip consistency could address occasional large errors. 

Our proof-of-concept demonstration illustrates how estimated posture can provide real-time, model-derived shoulder-strain feedback during robot-aided physiotherapy, and how the validated estimation noise affects this application. With sufficient knowledge of shoulder anatomy and real-time visual feedback, one of the authors could effectively adjust their arm null space to reduce the estimated shoulder strain. However, visual feedback alone might not be a sufficient tool for a naive patient. Our team's continued work aims to provide robot guidance and biomechanics-aware control, which could also make the demonstrated robot-aided physiotherapy accessible to non-expert patients and therapists. 


\section{Conclusion}
This study proposed and evaluated an RGB-D camera–robot fusion framework with an extended Kalman filter (EKF) for 7-DoF arm-state estimation during physical human–robot interaction. Matched EKF ablations showed that combining visual and robot measurements improved joint angle agreement with the motion-capture reference, while joint velocity errors remained similar to those of the robot-only estimator. The comparison between the assumed and participant-specific grip transformation revealed substantial sensitivity in distal joint estimates, highlighting the importance of modeling human–robot coupling for human pose estimation. As a proof of concept, we further demonstrated how such posture estimation, when combined with biomechanical modeling, can provide real-time visual feedback on the estimated shoulder strain. Overall, this work represents a step toward the proposed "walk-up-and-use" objective for more accessible robot-aided physiotherapy.

\bibliographystyle{IEEEtran}
\bibliography{IEEEexample}

\end{document}